%% file: emnlp2022.tex
\pdfoutput=1

\documentclass[11pt]{article}

\usepackage{EMNLP2022}

\usepackage{times}
\usepackage{latexsym}

\usepackage[T1]{fontenc}

\usepackage[utf8]{inputenc}

\usepackage{microtype}

\usepackage{inconsolata}
\usepackage{times}
\usepackage{latexsym}
\usepackage{hyperref}
\usepackage{microtype}
\usepackage{amsmath,amssymb}
\usepackage{epsfig,graphicx,subfigure,caption}
\usepackage{algpseudocode}
\usepackage[normalem]{ulem}
\usepackage{amsmath}
\usepackage[linesnumbered,algoruled,boxed,noend]{algorithm2e}
\usepackage{xcolor}
\usepackage{color, colortbl}
\usepackage{multirow,booktabs, hhline}
\usepackage{amsmath, bm}
\usepackage[linesnumbered,algoruled,boxed,noend]{algorithm2e}
\usepackage{wrapfig}
\usepackage{url}

\newcommand{\secref}[1]{\S\ref{#1}}

\newcommand{\eg}{\textit{e.g., }}
\newcommand{\ie}{\textit{i.e., }}

\newcommand{\xmdemoji}{\raisebox{-2pt}{\includegraphics[width=0.9em]{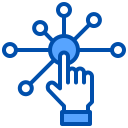}}}

\newcommand{\method}{\textbf{XMD}\xmdemoji}
\newcommand{\methodsp}{\textbf{XMD}\xmdemoji\xspace}
\DeclareUnicodeCharacter{FFFD}{}

\title{
\method: An End-to-End Framework for Interactive \\ Explanation-Based Debugging of NLP Models \\
\vspace{0mm}
\textcolor{black}{\footnotesize{\textit{Warning: This paper discusses and contains content that may be offensive or upsetting.}}}
}

\author{
Dong-Ho Lee\textsuperscript{1}\thanks{~~Both authors contributed equally.}~~~
Akshen Kadakia\textsuperscript{1}$^*$~
Brihi Joshi\textsuperscript{1}~
Aaron Chan\textsuperscript{1}~
Ziyi Liu\textsuperscript{1}~
Kiran Narahari\textsuperscript{1}~
\\
\textbf{Takashi Shibuya\textsuperscript{2}~
Ryosuke Mitani\textsuperscript{2}~
Toshiyuki Sekiya\textsuperscript{2}~
Jay Pujara\textsuperscript{1}~
Xiang Ren\textsuperscript{1}}
\\
\textsuperscript{1}Department of Computer Science, University of Southern California\\
\textsuperscript{2}R\&D Center, Sony Group Corporation\\
{\small \texttt{\{dongho.lee, akshenhe, brihijos, chanaaro, zliu2803, 
vnarahar, jpujara, xiangren\}@usc.edu}}\\
{\small \texttt{\{Takashi.Tak.Shibuya, Ryosuke.Mitani, Toshiyuki.Sekiya\}@sony.com}}
}

\begin{document}
\maketitle
\input{sections/0_abstract}
\input{sections/1_introduction}
\input{sections/2_overview}

\input{sections/3_framework}

\input{sections/4_implementation}
\input{sections/5_experiments}
\input{sections/6_related}

\input{sections/7_conclusion}

\bibliography{anthology,custom}
\bibliographystyle{acl_natbib}




\end{document}

%% file: sections/0_abstract.tex
\begin{abstract}
NLP models are susceptible to learning spurious biases (\ie bugs) that work on some datasets but do not properly reflect the underlying task.
Explanation-based model debugging aims to resolve spurious biases by showing human users explanations of model behavior, asking users to give feedback on the behavior, then using the feedback to update the model.
While existing model debugging methods have shown promise, their prototype-level implementations provide limited practical utility.
Thus, we propose \method: the first open-source, end-to-end framework for explanation-based model debugging.
Given task- or instance-level explanations, users can flexibly provide various forms of feedback via an intuitive, web-based UI.
After receiving user feedback, \methodsp automatically updates the model in real time, by regularizing the model so that its explanations align with the user feedback.
The new model can then be easily deployed into real-world applications via Hugging Face.
Using \method, we can improve the model's OOD performance on text classification tasks by up to 18\%.\footnote{Source code and project demonstration video are made publicly available at ~\url{http://inklab.usc.edu/xmd/}}


\end{abstract}

%% file: sections/1_introduction.tex
\section{Introduction}
\begin{figure}[t]
	\centering
	\vspace{2mm}
	\includegraphics[width=1\linewidth]{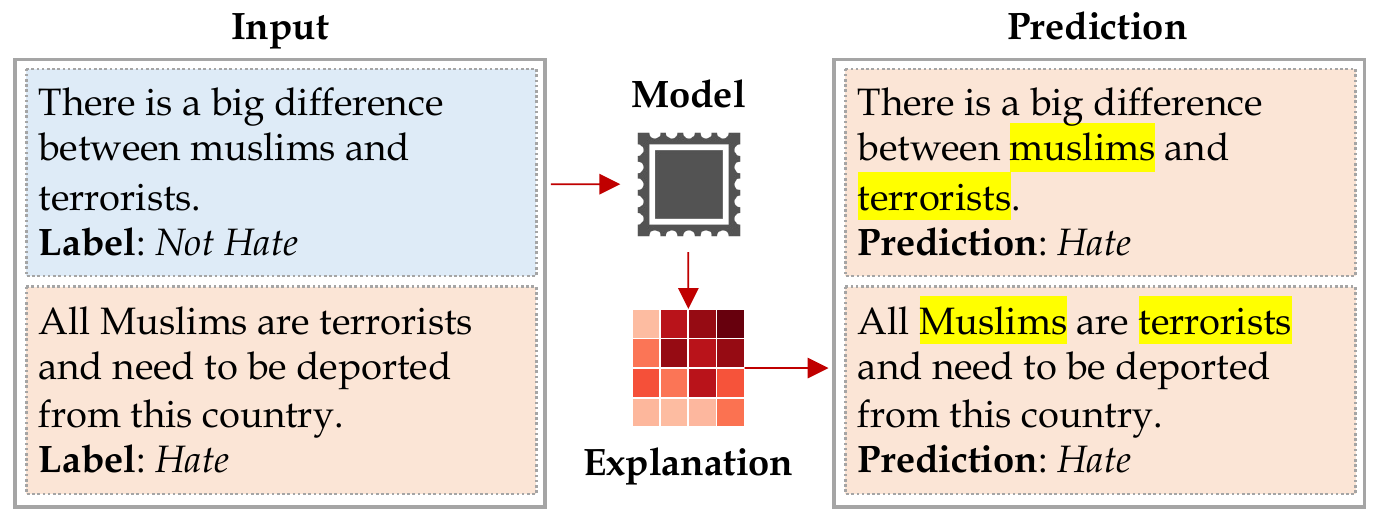}
	\caption{We make predictions on machine-generated examples~\cite{brown2020language, hartvigsen-etal-2022-toxigen} using BERT model fine-tuned on HateXplain~\cite{mathew2021hatexplain} and show its explanation using integrated gradients~\cite{sundararajan2017axiomatic}.
	It shows spurious correlation between a word \textit{muslims} and the label \textit{hate}.}
	\vspace{-10pt}
	\label{fig:motivation}
\end{figure}

Neural language models have achieved remarkable performance on a wide range of natural language processing (NLP) tasks~\cite{srivastava2022beyond}.
However, studies have shown that such NLP models are susceptible to learning spurious biases (\ie bugs) that work on specific datasets but do not properly reflect the underlying task~\cite{adebayo2020debugging, geirhos2020shortcut, du-etal-2021-towards, pmlr-v119-sagawa20a}.
For example, in hate speech detection, existing NLP models often associate certain group identifiers (\textit{e.g.}, \textit{black}, \textit{muslims}) with hate speech, regardless of how these words are actually used~\cite{kennedy-etal-2020-contextualizing} (Fig.~\ref{fig:motivation}).
This poses serious concerns about the usage of NLP models for high-stakes decision-making~\citep{bender2021dangers}.

In response, many methods have been proposed for debiasing either the model or the dataset.
Model debiasing can be done via techniques like instance reweighting \cite{schuster-etal-2019-towards}, confidence regularization \cite{utama2020mind}, and model ensembling \cite{he-etal-2019-unlearn, mahabadi2019simple, clark-etal-2019-dont}.
Dataset debiasing can be done via techniques like data augmentation~\cite{jia-liang-2017-adversarial, Kaushik2020Learning} and adversarial filtering~\cite{zellers-etal-2018-swag, le2020adversarial}.
However, these methods lack knowledge of which spurious biases actually impacted the model's decisions, which greatly limits their debiasing ability.


On the other hand, \textit{explanation-based model debugging} focuses on addressing spurious biases that actually influenced the given model's decision-making \cite{smith2020no, lertvittayakumjorn2021explanation, hartmann2022survey}.
In this paradigm, a human-in-the-loop (HITL) user is given explanations of the model's behavior \cite{sundararajan2017axiomatic, shrikumar2017learning} and asked to provide feedback about the behavior.
Then, the feedback is used to update the model, in order to correct any spurious biases detected via the user feedback.
While existing model debugging methods have shown promise \cite{idahl2021towards, lertvittayakumjorn-etal-2020-find, zylberajch-etal-2021-hildif, ribeiro2016should}, their prototype-level implementations provide limited end-to-end utility (\ie explanation generation, explanation visualization, user feedback collection, model updating, model deployment) for practical use cases.

Given the interactive nature of explanation-based model debugging, it is important to have a user-friendly framework for executing the full debugging pipeline. 
To achieve this, we propose the E\textbf{X}planation-Based NLP \textbf{M}odel \textbf{D}ebugger (\textbf{\method}).
Compared to prior works, \methodsp makes it simple for users to debug NLP models and gives users significant control over the debugging process (Fig. \ref{fig:overview}).
Given either task (model behavior over all instances) or instance (model behavior w.r.t. a given instance) explanations, users can flexibly provide various forms of feedback (\eg \textit{add} or \textit{remove} focus on a given token) through an easy-to-use, web-based user interface (UI).
To streamline user feedback collection, \method's UI presents intuitive visualizations of model explanations as well as the different options for adjusting model behavior (Fig. \ref{fig:localexpui}-\ref{fig:globalexpui}).
After receiving user feedback, \methodsp automatically updates the model in real time, by regularizing the model so that its explanations align with the user feedback \cite{joshi2022er}.
\methodsp also provides various algorithms for conducting model regularization.
The newly debugged model can then be downloaded and imported into real-world applications via Hugging Face \cite{wolf-etal-2020-transformers}.
To the best of our knowledge, \methodsp is the first open-source, end-to-end framework for explanation-based model debugging. 
We summarize our contributions as follows:

\noindent
\textbullet\ \hspace{0.2mm} \textbf{End-to-End Model Debugging}:
\methodsp packages the entire model debugging pipeline (\ie explanation generation, explanation visualization, user feedback collection, model updating, model deployment) as a unified system. \methodsp is agnostic to the explanation method, user feedback type, or model regularization method.
\method can improve models' out-of-distribution (OOD) performance on text classification tasks (\eg hate speech detection, sentiment analysis) by up to 18\%.

\noindent
\textbullet\ \hspace{0.2mm} \textbf{Intuitive UI}:
\method's point-and-click UI makes it easy for non-experts to understand model explanations and give feedback on model behavior.

\noindent
\textbullet\ \hspace{0.2mm} \textbf{Easy Model Deployment}:
Given user feedback, \methodsp automatically updates the model in real time.
Users can easily deploy debugged models into real-world applications via Hugging Face.

%% file: sections/2_overview.tex
\section{Framework Overview}

\begin{figure}[t]
	\centering
	\vspace*{-.1in}
	\includegraphics[width=1\linewidth]{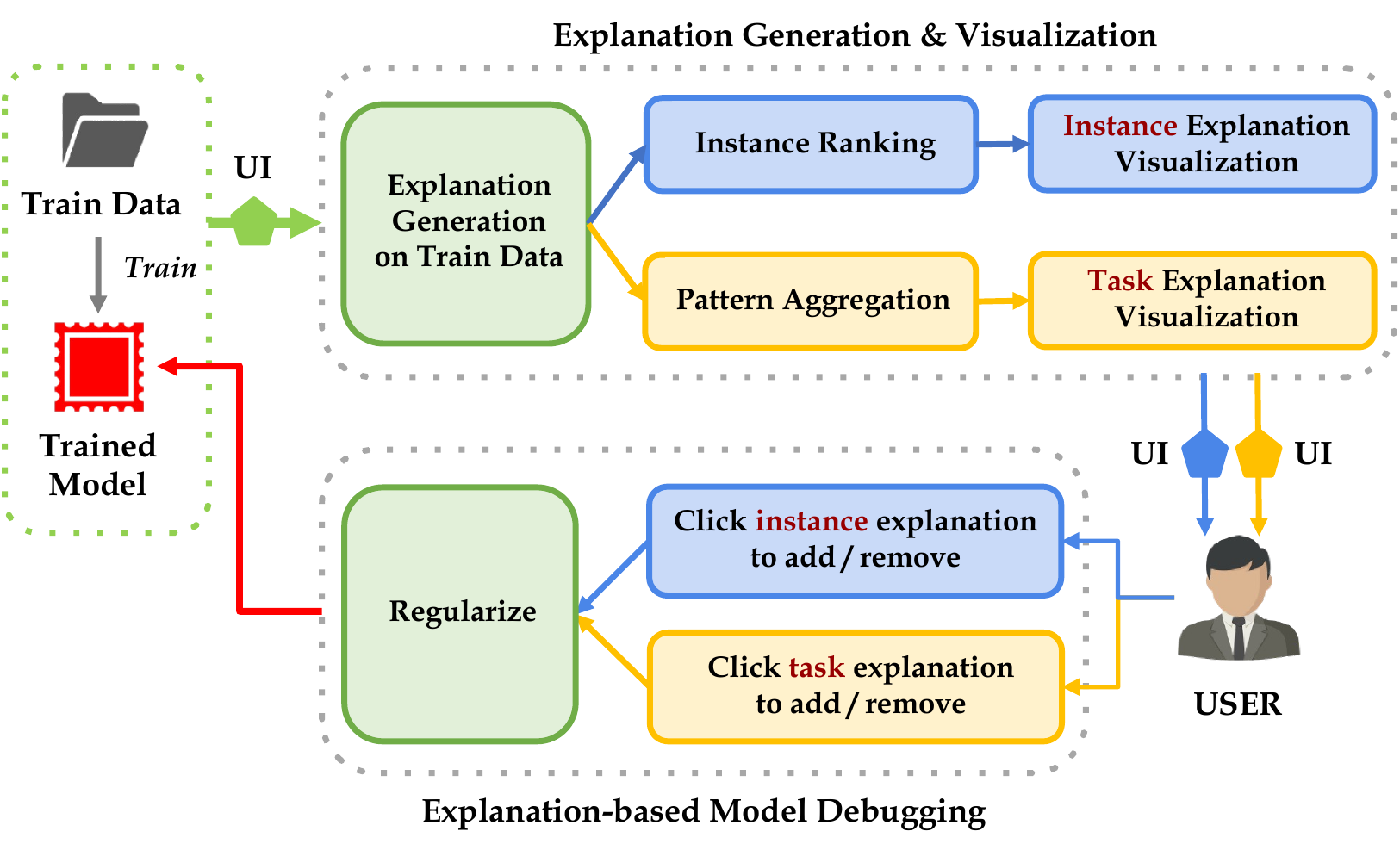}
	\caption{\textbf{System Architecture}}
	\vspace{-10pt}
	\label{fig:overview}
\end{figure}


As shown in Figure~\ref{fig:overview}, our framework consists of three main components: \textit{Explanation Generation}; \textit{Explanation Visualization}; and \textit{Explanation-based Model Debugging}.
Here, the explanation generation and debugging process are done on the back-end while visualizing explanations and capturing human feedback on them are done on front-end UI.

\paragraph{Explanation Generation (\secref{ssec:eg})}
Humans first input the train data and the model trained on the train data into the framework.
On the backend, our framework uses a heuristic post-hoc explanation approach on the model to generate rationales for the train data.

\paragraph{Explanation Visualization (\secref{ssec:ev})}
The framework visualizes the generated rationales through UI in two different ways: an instance explanation, which shows the explanation for each train instance, and a task explanation, which shows words according to their importance to the prediction.

\paragraph{Explanation-Based Model Debugging (\secref{ssec:ebhd})}
The human then decides whether to select words for each instance (Instance Explanation) or words that apply to all instances (Task Explanation) to increase or decrease the word importance.
When a human clicks a few words to debug and then decides to end the debugging process, the framework retrains the model with a regularization approach and makes the debugged model downloadable.

%% file: sections/3_framework.tex
\section{\methodsp Framework}

In this section, we present each module of \methodsp in processing order.
To start the process, the user needs to place a training dataset $\mathcal{D}_T$ and a classification model $\mathcal{M}$ that is trained on $\mathcal{D}_T$.

\subsection{Explanation Generation}
\label{ssec:eg}

Our explanation generation module outputs rationales from $\mathcal{M}$.
For each instance $\mathbf{x}\in\mathcal{D}_T$, $\mathcal{M}$ generates rationales $\phi({\mathbf{x}}) = [\phi({\mathbf{w}_1}), \phi({\mathbf{w}_2}), \dots, \phi({\mathbf{w}_n})]$ where $\mathbf{w}_i$ denotes the i-th token in the sentence.
Each importance score $\phi({\mathbf{w}_i})$ has a score with regard to all the classes.
Our module is exploiting $\phi^{p}({\mathbf{w}_i})$ which the importance score is attributed to model predicted label $p$.
Here, we exploit heuristic methods that assign importance scores $\phi$ based on gradient changes in $\mathcal{M}$~\cite{shrikumar2017learning, sundararajan2017axiomatic}.



\subsection{Explanation Visualization}
\label{ssec:ev}

Our framework supports visualizing the generated rationale in two different forms, \textit{instance} and \textit{task} explanations.
Instance explanations display word importance scores for model predictions for each train instance, while task explanations aggregate and rank words according to their importance to the predicted label.
In this section, we first present a UI for visualizing and capturing human feedback for instance explanations and then a UI for task explanations.

\paragraph{Instance Explanation}

\begin{figure}[!t]
	\centering
	\setlength\belowcaptionskip{-1\baselineskip}
	\includegraphics[width=\linewidth]{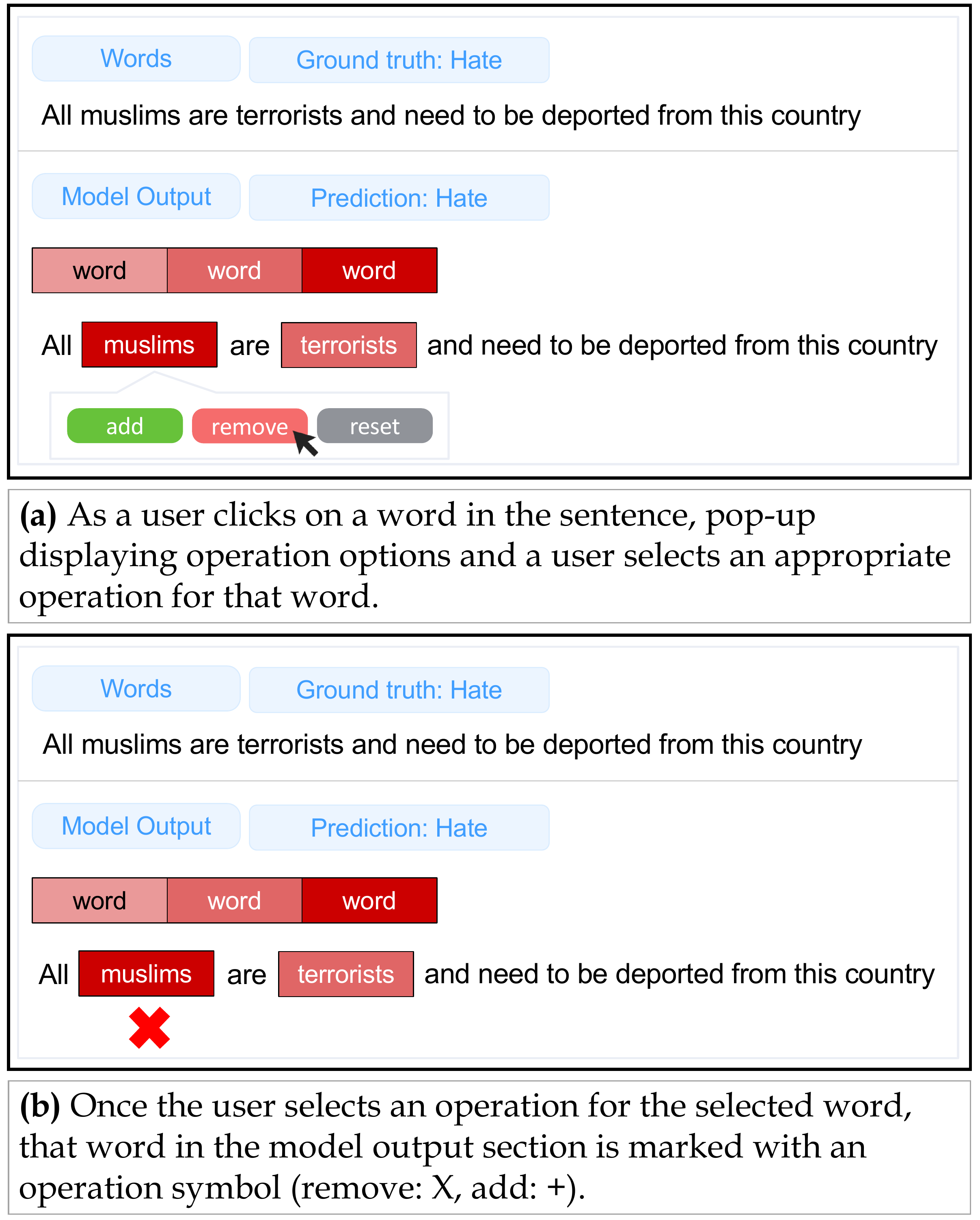}
	\caption{The workflow to provide human feedback on \textbf{instance explanations}. Humans provide explanations (\textit{remove ``muslims''}) for the ground truth label (\textit{hate}).}
	\label{fig:localexpui}
\end{figure}

Figure~\ref{fig:localexpui} illustrates how our framework visualizes instance explanations and captures human feedback on them.
First, the trained model makes a prediction and generates explanations for one of the train instances that the model correctly predicts: ``All muslims are terrorists and need to be deported from this country''.
The reason why we present only the instances that the model correctly predicts is that we are asking users to provide feedback for the ground truth label and comparing it with $\phi^{p}({\mathbf{w}_i})$ which the importance score is attributed to the model predicted label $p$.
If $p$ is not equal to the ground truth label, the human feedback would act as a source of incorrect prediction.

Next, the user is presented with the sentence and its ground truth label on the upper deck (Words Section), and the sentence with highlighted rationales and its predicted label on the lower deck (Model Output Section).
Then, the user can choose to select words to decrease or increase its importance toward the ground truth label (Figure~\ref{fig:localexpui} (a)).
If the user clicks the word (\textit{muslims}) that the model is focusing on to predict \textit{hate}, a small pop-up displaying buttons for operation options (\textit{i.e.}, \textit{add}, \textit{remove} and \textit{reset}) appear.
Once the user selects a desired operation (\textit{remove}) for the selected word (\textit{muslims}) that is not a right reason for \textit{hate}, that word in the model output section is marked with operation symbol ('X' for \textit{remove}, '+' for \textit{add} -- Figure~\ref{fig:localexpui} (b)).
The user may cancel their decision to operation for the word by clicking \textit{reset} in the pop-up.

\paragraph{Task Explanation}

\begin{figure}[!t]
	\centering
	\setlength\belowcaptionskip{-1\baselineskip}
	\includegraphics[width=\linewidth]{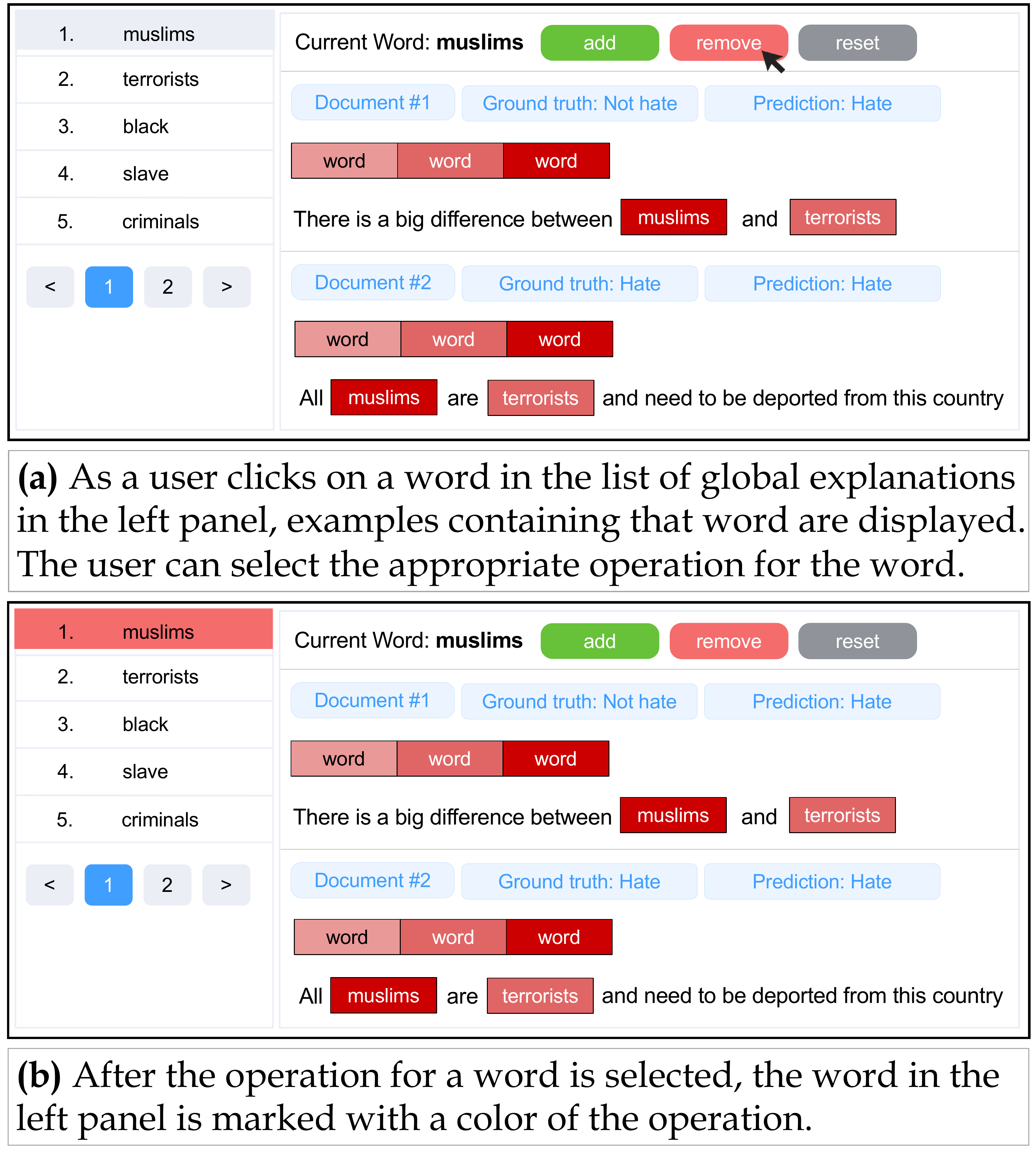}
	\caption{The workflow to provide human feedback on \textbf{task explanations}. Humans remove the word (\textit{muslims}) that should not be conditioned on any labels (i.e., \textit{hate}, \textit{not hate}).}
	\label{fig:globalexpui}
\end{figure}

Figure~\ref{fig:globalexpui} illustrates how our framework visualizes task explanations and captures human feedback on them.
First, task explanations are presented in list format on the left panel in descending order of its importance (Figure~\ref{fig:globalexpui} (a)).
Here, the importance is a score averaged by the word importance score of all examples containing that word.
As user clicks on a word in the list, all the examples containing that word are displayed.
The user can then choose to two different operations (\textit{remove} and \textit{add}).
If user clicks \textit{remove} for the word (\textit{muslims}) that should not be conditioned on any label (both \textit{hate} and \textit{not hate}), the model will consider it as an unimportant word in all cases.
Here, we don't need to consider whether the prediction is correct or not since the word is not important for all the cases (Figure~\ref{fig:globalexpui} (a)).
If user clicks \textit{add} for the word that should be useful for the correct prediction, the model will consider it as an important word for the ground truth label.
Here, we consider it as an important word only for the correct prediction.
After the operation for a word is selected, the word in the left panel is marked with a color of that operation (red for \textit{remove} and green for \textit{add}).

\subsection{Explanation-based Model Debugging}
\label{ssec:ebhd}

\begin{figure}[t]
	\centering
	\includegraphics[width=1\linewidth]{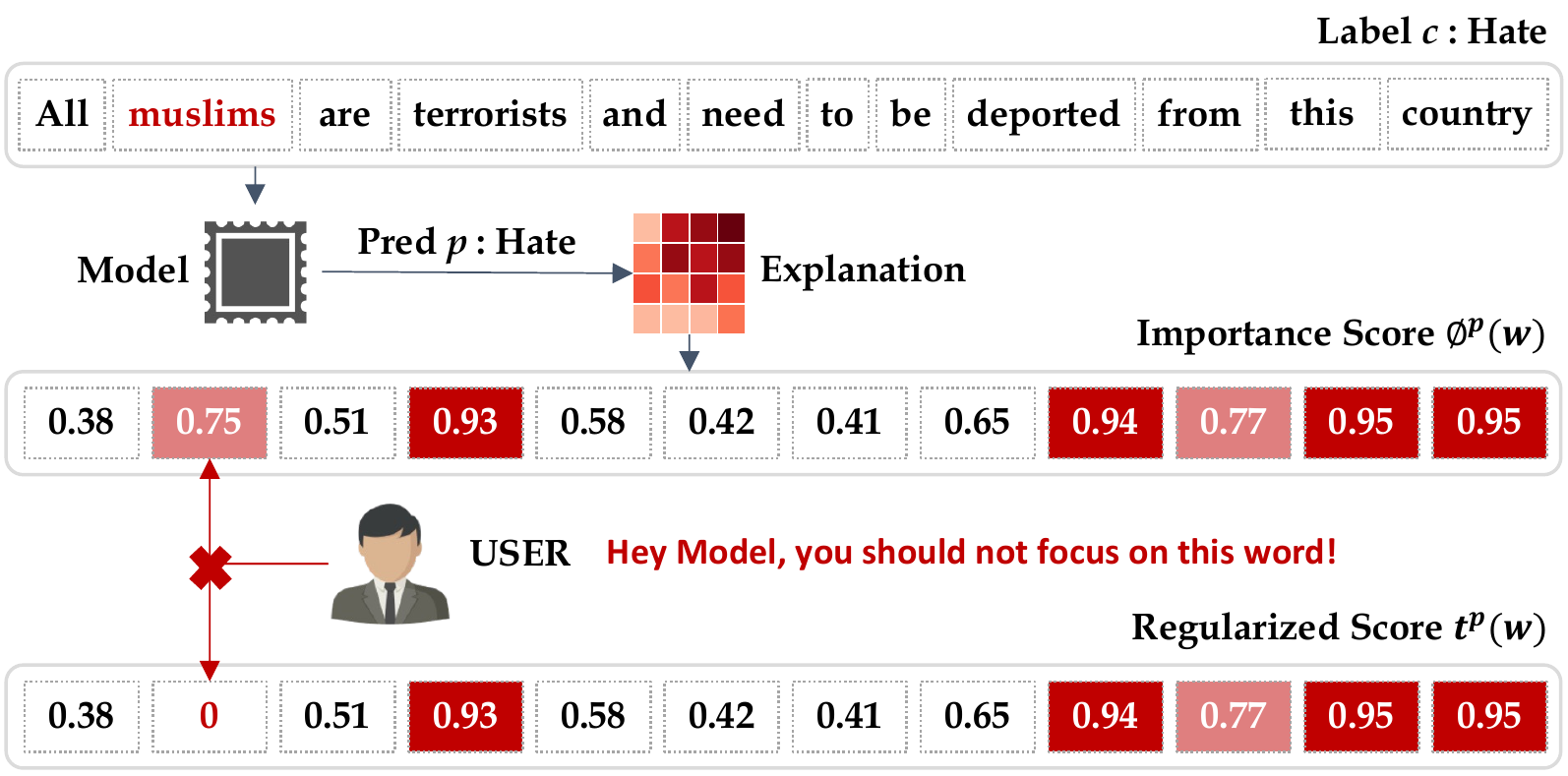}
	\caption{\textbf{Instance Explanation-based Model Debugging.} Trained model generates explanations in a form of word importance score $\phi^{p}(w)$ towards prediction label $\mathbf{p}$. User selects words to \textit{add} or \textit{remove} based on $\phi^{p}(w)$. The regularization score $t^{p}(w)$ for the selected words to be removed are 0 while selected words to add are 1.}
	\label{fig:localexpreg}
\end{figure}

Our explanation-based model debugging module is based on explanation regularization (ER) which regularizes model to produce rationales that align to human rationales~\cite{zaidan2008modeling, ross2017right, liu2019incorporating, ghaeini2019saliency, kennedy2020contextualizing, rieger2020interpretations, lin2020triggerner, huang2021exploring, joshi2022er}.
Existing works require the human to annotate rationales for each training instance or apply task-level human priors (\textit{e.g.}, task-specific lexicons) across all training instances before training.
Despite its effectiveness, the regularized model may not be fully free of hidden biased patterns.
To catch all the hidden biased patterns, our framework asks the human to provide binary feedback (\textit{i.e.}, click to add or remove) given the current model explanations and use them to regularize the model.
Here we ask the human to provide feedback to the model in order to output the ``\textbf{correct prediction}''.

\begin{figure}[t]
	\centering
	\includegraphics[width=1\linewidth]{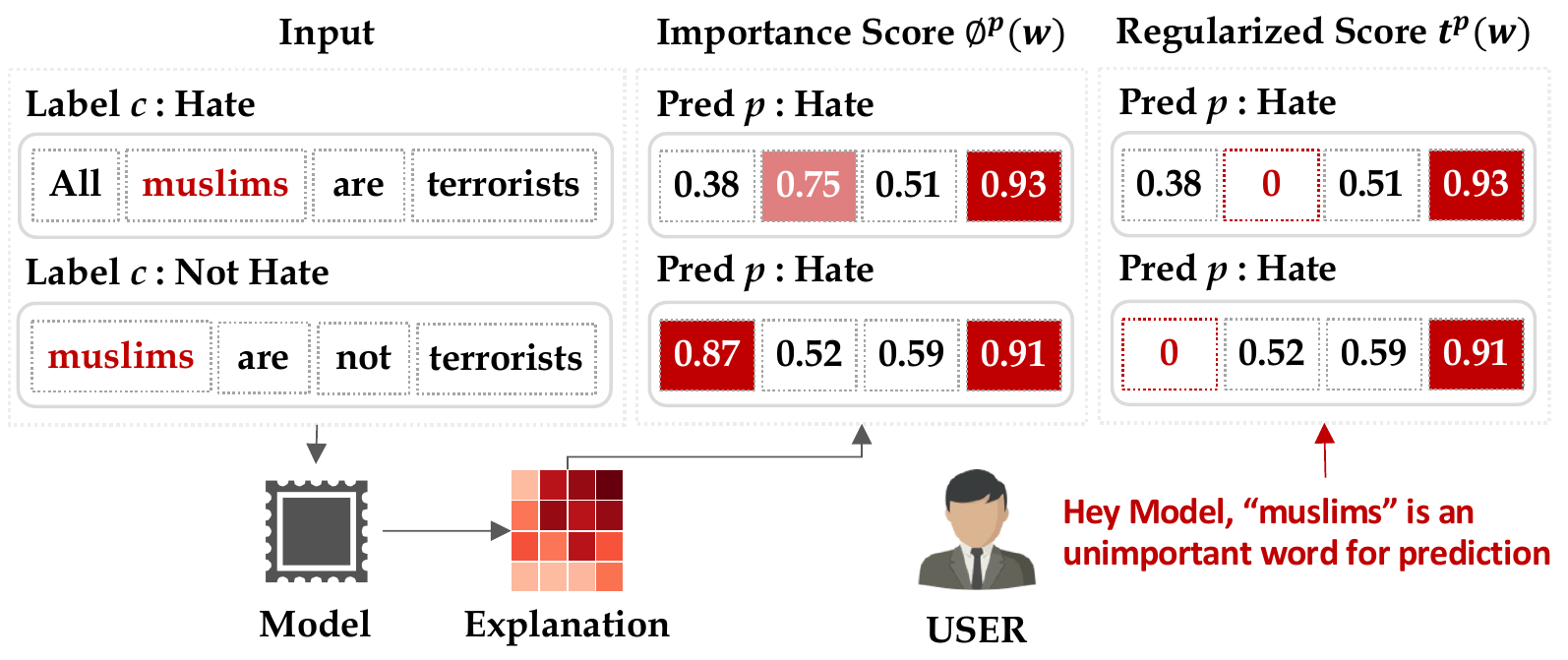}
	\caption{\textbf{Task Explanation-based Model Debugging.} Trained model generates explanations in a form of word importance score $\phi^{p}(w)$ towards prediction label $\mathbf{p}$.
	As user selects a word to ignore for prediction, the regularization score $t^{p}(w)$ for the selected word in all the examples that contain that word becomes 0.
	}
	\label{fig:globalexpreg}
\end{figure}

For the instance explanation, as shown in Figure~\ref{fig:localexpreg}, the trained model $\mathcal{M}$ generates rationales $\phi^{p}({\mathbf{x}}) = [\phi^{p}({w_1}), \phi^{p}({w_2}), \dots, \phi^{p}({w_n})]$, where $\phi^{p}({w_i})$ denotes the importance score of i-th token in the sentence $\mathbf{x}$ towards model predicted label $p$.
As the user selects a word (\textit{muslims}) that is spuriously correlated with the correct prediction $p$ (\textit{hate}), the regularized score $t^{p}({w_i})$ where $i$ is a user-selected word index ($w_2=$ \textit{muslims}) becomes 0 (See Figure~\ref{fig:localexpreg}).
For the task explanation, we aggregate words based on its score averaged by the word importance score of examples containing that word, and present them in a descending order.
When the user clicks a word $w$ (\textit{muslims}) to decrease its importance, then regularized score $t^{p}({w})$ where $w$ is user-selected word ($w=$ \textit{muslims}) for all the examples become 0  (See Figure~\ref{fig:globalexpreg}).

After the click process, the user can start the debugging process based on the examples labeled so far.
Here, the learning objective for re-training the model $\mathcal{M}$ is $\mathcal{L}=\mathcal{L}_{\text {task }}+\mathcal{L}_{\mathrm{ER}}$, where  $\mathcal{L}_{\text {task}}$ is a cross-entropy loss for traditional sequence classification tasks and $\mathcal{L}_{\mathrm{ER}}$ is an explanation regularization loss which minimizes the distance between $\phi^{p}({w})$ and $t^{p}({w_i})$~\cite{joshi2022er}.
In this framework, we support two different regularization loss: Mean Squared Error (MSE)~\cite{liu-avci-2019-incorporating, kennedy-etal-2020-contextualizing, ross2017right}, Mean Absolute Error (MAE)~\cite{rieger2020interpretations}.

%% file: sections/4_implementation.tex
\section{Implementation Details}

To start \methodsp, users should input the trained model following Hugging Face model structure~\cite{wolf-etal-2020-transformers}.
After users input the train data and the model, our framework uses Captum~\cite{kokhlikyan2020captum} to generate explanation.
For visualizing the explanation and capturing the human feedback, we implement UI using Vue.js~\footnote{https://vuejs.org/}.
Here, we re-use UI components from LEAN-LIFE, an explanation-based annotation framework~\cite{lee-etal-2020-lean}, for capturing human feedback.
To train the model with ER, we use PyTorch~\cite{NEURIPS2019_9015} and Huggingface~\cite{wolf-etal-2020-transformers}.

%% file: sections/5_experiments.tex
\section{Experiments}

We conduct extensive experiments investigating how our debugging process affects the performance on in-distributed (ID) and out-of-distribution (OOD) data, and the model explanation.
Here, we present experimental results on sentiment analysis for the instance explanation and hate speech detection for the task explanation.
For base model, we use BigBird-Base~\cite{zaheer2020big}.

\paragraph{Tasks and Datasets}
For sentiment analysis, we exploit SST~\cite{socher2013recursive} as the ID dataset, and Yelp (restaurant reviews) \citep{zhangCharacterlevelConvolutionalNetworks2015}, Amazon (product reviews) \citep{mcauley2013hidden} and Movies (movie reviews) \cite{zaidan2008modeling, deyoung2019eraser} as OOD datasets.
To simulate human feedback for the instance explanation, we leverage ground truth rationales for SST~\cite{carton2020evaluating} as human feedback.
For hate speech detection, we use STF~\cite{de-gibert-etal-2018-hate} as the ID dataset, and 
HatEval~\cite{barbieri-etal-2020-tweeteval}, Gab Hate Corpus (GHC)~\cite{kennedyghc} and Latent Hatred~\cite{elsherief-etal-2021-latent} for OOD datasets.
To simulate human feedback for the task explanations, we leverage group identifiers (\textit{e.g.}, \textit{black}, \textit{muslims})~\cite{kennedy-etal-2020-contextualizing} as words that need to be discarded for determining whether the instance is hate or not.

\paragraph{ID/OOD Performance}
Table~\ref{tab:localexp} shows the performance on ID and OOD when regularize on correct predictions using its instance explanation.
We see that our framework helps model to not only do much better on ID data, but also generalize well to OOD data.
For task explanation, we present performance by regularizing on correct and incorrect prediction and all the instances regardless of prediction.
Table~\ref{tab:globalexp} presents the performance with \textit{remove} operations for task explanations (\textit{i.e.}, group identifiers) for incorrect predictions, correct predictions, and for all instances, respectively.
We observe that our framework helps model not to focus on the words that should not be conditioned on any label and lead to performance enhancement on both ID and OOD data.

\begin{table}[t!]
\vspace{0.1cm}
	\centering
	\scalebox{0.68
	}{
		\begin{tabular}{cccccc}
			\toprule
			\multirow{3}{*}{\textbf{Regularize}} & \multirow{3}{*}{\textbf{ER Loss}} & \multicolumn{4}{c}{\textbf{Sentiment Analysis}} \\
			\cmidrule(lr){3-6}
			& & In-distribution & \multicolumn{3}{c}{Out-of-Distribution} \\
			\cmidrule(lr){3-3} \cmidrule(lr){4-6}
            & & SST & Amazon & Yelp & Movies \\
			\midrule
			None & None & 93.4 & \underline{89.1} & 89.0 & 82.0 \\
			\midrule
			Correct & MSE & \bf 94.7 & 88.4 & \underline{91.8} & \bf 94.5\\
			 & MAE & \underline{94.0} & \bf 92.3 & \bf 94.4 & \underline{94.0} \\
			\bottomrule
		\end{tabular}
	} 
	\caption{\textbf{Instance Explanation} ID/OOD Performance (Accuracy). Best models are bold and second best ones are underlined within each metric.}
	\label{tab:localexp}
	\vspace{-0.4cm}
\end{table}

\begin{table}[t!]
\vspace{0.1cm}
	\centering
	\scalebox{0.68
	}{
		\begin{tabular}{cccccc}
			\toprule
			\multirow{3}{*}{\textbf{Regularize}} & \multirow{3}{*}{\textbf{ER Loss}} & \multicolumn{4}{c}{\textbf{Hate Speech Analysis}} \\
			\cmidrule(lr){3-6}
			& & In-distribution & \multicolumn{3}{c}{Out-of-Distribution} \\
			\cmidrule(lr){3-3} \cmidrule(lr){4-6}
            & & STF & HatEval & GHC & Latent \\
			\midrule
			None & None & 89.5 & 88.2 & 64.5 & 67.2 \\
			\midrule
			Correct & MSE & 89.2 & \bf 90.1 & 62.3 & 67.9\\
		            & MAE	 & 89.1 & \bf 90.1 & 59.3 & 64.9 \\
			\midrule
			Incorrect & MSE & 88.9 & 86.3 & \bf 67.9 & \bf 70.3 \\
			 & MAE & 89.3 & \underline{88.8} & 64.2 & 67.6 \\
 			\midrule
			ALL & MSE & \bf 90.0 & 88.4 & 63.8 & 67.0 \\
			 & MAE & \underline{89.7} & 86.9 & \underline{66.5} & \underline{70.2} \\
			\bottomrule
		\end{tabular}
	} 
	\caption{\textbf{Task Explanation} ID/OOD Performance (Accuracy). Best models are bold and second best ones are underlined within each metric.}
	\label{tab:globalexp}
	\vspace{-0.4cm}
\end{table}

\paragraph{Efficiency}

\begin{figure*}
    \centering
    \includegraphics[width=0.97\linewidth]{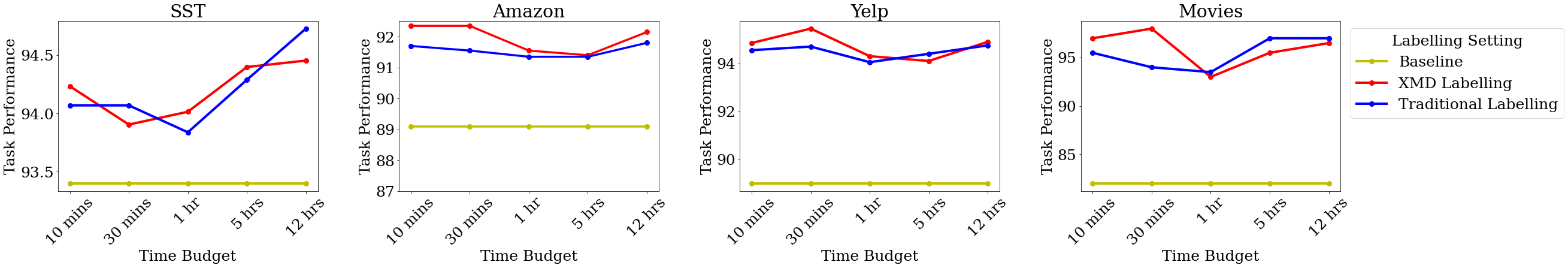}
    \vspace{-0.1cm}
    \caption{\textbf{Time Efficiency:} This simulation assumes two annotators annotating instances in parallel with a strict time budget, using the traditional labelling ($\sim110$s/instance) or \methodsp labelling ($\sim60$s/instance) methods.}
    \label{fig:exp:efficiency}
    \vspace{-0.3cm}
\end{figure*}

To quantify the advantage that \methodsp provides, we compare the time taken to annotate instances using \methodsp versus traditional labelling for instance explanations. While \methodsp requires humans to interact with a trained model and decrease or increase importance scores of words, traditional labelling is not model-in-the-loop in nature, and requires users to directly annotate binary importance scores to words in the instance \cite{deyoung2019eraser, carton2020evaluating}.
We ask two graduate students to annotate $50$ instances, using the traditional and the \methodsp labelling methods. For both of these labelling settings, we ensure that there is no overlap between the instances, so as to avoid familiarity and record the time taken to annotate each instances. Upon aggregating across all instances and both annotators, it is found that one instance takes $\sim60$ seconds and $\sim110$ seconds to be annotated using the framework and the traditional labelling method respectively. Using this time estimate, we simulate the time-efficiency of these two labelling methods with varying amounts of time budgets for annotations. Figure \ref{fig:exp:efficiency} presents our results for this experiment. We note that although both labelling methods outperforms the baseline of no explanation annotation, using \methodsp is particularly helpful when the time budget given is limited (< $1$ hour), especially in the OOD setting (Amazon, Yelp, Movies datasets).




%% file: sections/6_related.tex
\section{Related Work}

\paragraph{Spurious Bias Mitigation}
Recent studies have explored mitigating spurious biases in NLP models.
One of the research lines is a dataset debiasing such as adversarial filtering~\cite{zellers-etal-2018-swag, le2020adversarial} or data augmentation using adversarial data~\cite{jia-liang-2017-adversarial} and counterfactual data~\cite{Kaushik2020Learning}.
However, creating such datapoints are challenging since they require an exhaustive understanding of the preconceived notions that may cause such spurious biases and the collecting cost is expensive.
Another line of research is robust learning techniques such as instance reweighting \cite{schuster-etal-2019-towards}, confidence regularization \cite{utama2020mind}, and model ensembling \cite{he-etal-2019-unlearn, mahabadi2019simple, clark-etal-2019-dont}.

\paragraph{Explanation-Based Model Debugging}
Many works have explored explanation-based debugging of NLP models, mainly differing in how model behavior is explained, how HITL feedback is provided, and how the model is updated \cite{lertvittayakumjorn2021explanation, hartmann2022survey, balkir2022challenges}.
Model behavior can be explained using instance \cite{idahl2021towards, koh2017understanding, ribeiro2016should} or task \cite{lertvittayakumjorn-etal-2020-find, ribeiro2018semantically} explanations, typically via feature importance scores.
HITL feedback can be provided by modifying the explanation's feature importance scores \cite{kulesza2009fixing, kulesza2015principles, zylberajch-etal-2021-hildif} or deciding the relevance of high-scoring features \cite{lu2022rationale, kulesza2010explanatory, ribeiro2016should, teso2019explanatory}.
The model can be updated by directly adjusting the model parameters \cite{kulesza2009fixing, kulesza2015principles, smith2020no}, improving the training data \cite{koh2017understanding, ribeiro2016should, teso2019explanatory}, or influencing the training process \cite{yao2021refining, cho2019explanatory, stumpf2009interacting}.
In particular, explanation regularization (ER) influences the training process so that the model's explanations align with human explanations \cite{joshi2022er, ross2017right, kennedy2020contextualizing, rieger2020interpretations, liu2019incorporating, chan2022unirex}.

Our \methodsp system is agnostic to the choice of explanation method or HITL feedback type, while updating the model via ER.
Compared to prior works, \methodsp gives users more control over the interactive model debugging process.
Given either global or local explanations, users can flexibly provide various forms of feedback via an intuitive, web-based UI.
After receiving user feedback, \methodsp automatically updates the model in real time.
The debugged model can then be downloaded and imported into real-world applications via Hugging Face \cite{wolf-etal-2020-transformers}.

%% file: sections/7_conclusion.tex
\section{Conclusion}

In this paper, we propose an open-source and web-based explanation-based NLP Model Debugging framework \methodsp that allows user to provide various forms of feedback on model explanation.
This debugging process guides the model to make predictions with the correct reason and lead to significant improvement on model generalizability.
We hope that \methodsp will make it easier for researchers and practitioners to catch spurious correlations in the model and debug them efficiently.